\documentclass[final]{cvpr}
\usepackage{times}
\usepackage{epsfig}
\usepackage{graphicx}
\usepackage{amsmath}
\usepackage{amssymb}
\usepackage[ruled,linesnumbered]{algorithm2e}

\usepackage{ bbold }
\usepackage[utf8]{inputenc} 
\usepackage[T1]{fontenc}    
\usepackage{multirow}
\usepackage{booktabs}

\usepackage{float}
\usepackage{caption}
\usepackage{subcaption}
\usepackage{soul}

\usepackage[font=small,skip=0pt]{subcaption}
\usepackage[font=small,skip=8pt]{caption}
\usepackage{rotating}
\usepackage{balance}

\usepackage{amsmath}

\DeclareMathOperator*{\argmin}{arg\,min}

\newlength\myindent
\DeclareMathAlphabet\mathbfcal{OMS}{cmsy}{b}{n}

\newcommand{\ilkd}{CCIL }
\newcommand{\ilkdsd}{CCIL-SD }
\newcommand{\cfr}{$\mathbfcal{R}_{\phi}$} 

\usepackage{xcolor, colortbl}
\colorlet{dgreen}{green!50!black}
\colorlet{dred}{red!50!black}
\def\yo{\ensuremath\checkmark}
\def\byo{\pmb{\ensuremath\checkmark}}

\usepackage[pagebackref=true,breaklinks=true,colorlinks,bookmarks=false]{hyperref}

\begin{document}

\title{Essentials for Class Incremental Learning}

\author{Sudhanshu Mittal \qquad Silvio Galesso   \qquad Thomas Brox \\  University of Freiburg, Germany}

\maketitle

\begin{abstract}
Contemporary neural networks are limited in their ability to learn from evolving streams of training data. When trained sequentially on new or evolving tasks, their accuracy drops sharply, making them unsuitable for many real-world applications. 
In this work, we shed light on the causes of this well known yet unsolved phenomenon - often referred to as \textit{catastrophic forgetting} - in a class-incremental setup.
We show that a combination of simple components and a loss that balances intra-task and inter-task learning can already resolve forgetting to the same extent as more complex measures proposed in literature.
Moreover, we identify poor quality of the learned representation as another reason for catastrophic forgetting in class-IL. We show that performance is correlated with secondary class information (\textit{dark knowledge}) learned by the model and it can be improved by an appropriate regularizer. With these lessons learned, class-incremental learning results on CIFAR-100 and ImageNet improve over the state-of-the-art by a large margin, while keeping the approach simple. 
\end{abstract}

\section{Introduction}


The ability to learn from continuously evolving data is important for many real-world applications.
Latest machine learning models, especially artificial neural networks, have shown great ability to learn the task at hand, but when confronted with a new task, they tend to override the previous concepts. Deep networks suffer heavily from this catastrophic forgetting~\cite{mccloskey:catastrophic} when trained with a sequence of tasks, impeding continual or lifelong learning.

In this work, we focus on class-incremental learning (class-IL) \cite{icarl}. It is one of the three scenarios of continual learning as described in \cite{1904.07734}, where the objective is to learn a unified classifier over incrementally occurring sets of classes. Since all the incremental data cannot be retained for unified training, the major challenge is to avoid forgetting previous classes while learning new ones.  


The three crucial components of a class-IL algorithm include a memory buffer to store few exemplars from old classes, a forgetting constraint to keep previous knowledge while learning new tasks, and a learning system that balances old and new classes. 
Although several methods have been proposed to address each of these components, there is not yet a common understanding of best practices.

Prabhu~\etal~\cite{gdumb} provides an overview over current state of continual learning methods for classification. It shows that a simple greedy balanced sampler-based approach (GDumb) can outperform various specialized formulations in most of the continual learning settings, however, it finds class-IL particularly challenging. In this work, we propose a complementary approach to~\cite{gdumb} for class-IL, where softmax outputs are masked appropriately with data balancing to outperform previous sophisticated approaches.

\paragraph{Contributions.} We propose a compositional class-IL (CCIL) model that isolates the underlying reasons for catastrophic forgetting in class-IL and combines the most simple and most effective components to build a robust base model.
It employs plain knowledge distillation \cite{kd} as a forgetting constraint and selects exemplar samples simply randomly. 
For the loss evaluation, we propose important changes in the output normalization. 
This base model already exceeds state-of-the-art results by a good margin.

In addition, we study the influence of the learned representation's properties on forgetting and show that the degree of feature specialization (overfitting) correlates with the degree of forgetting. We study some common regularization techniques and show that only those that keep, or even improve, the so-called secondary class information -- also referred as dark knowledge by \cite{kd} -- have a positive influence on class-incremental learning, whereas others make things much worse. The source code of this paper is available \footnote{Source code: \url{https://github.com/sud0301/essentials_for_CIL}}.

\section{Related Work}
\label{sec:related_work}



iCaRL was the first approach that formally introduced the class-IL problem~\cite{icarl}. iCaRL is a decoupled approach for feature representation learning and classifier learning. It alleviates catastrophic forgetting via knowledge distillation and a replay-based approach. Later Castro \etal ~\cite{eeil} extended it to an end-to-end learning model based on a combination of distillation and cross-entropy loss to show improved results over iCaRL.
Successive works usually dedicated their contribution to one of the three components in class-IL.

\paragraph{Exemplar selection:}
Replay-based approaches have been shown to be quite effective in mitigating catastrophic forgetting. Typically, a memory buffer is allocated to store exemplar samples of old classes, which are replayed while learning a new task to mitigate forgetting. Many works~\cite{eeil, lucir, icarl, bic} use herding heuristics~\cite{herding} for exemplar selection. Herding selects and retains samples closest to the mean sample for each class.
Liu~\etal~\cite{mnemonics} parameterized the exemplars to optimize them jointly with the model. Iscen~\etal~\cite{memory_efficient} introduced a memory efficient approach to store feature descriptors instead of images. In our work, we simply sample from each class randomly to compile the exemplar set. 

\paragraph{Forgetting-constraint:}
Knowledge distillation (KD) was first introduced by Li~\etal~\cite{lwf} for multi-task incremental learning. Thereafter, various works~\cite{eeil, icarl, bic} have adopted it in class-IL to restore previous knowledge. Lately, several works have proposed new forgetting constraints with an objective to preserve the structure of old-class embeddings. Hou~\etal~\cite{lucir} proposed the usage of feature-level distillation by penalizing change is the feature representation from the old model. Yu~\etal~\cite{semantic_drift} utilized an embedding network to rectify the semantic drift, Tao~\etal~\cite{tpcil} proposed a Hebbian graph-based approach to retain the topology of the feature space. In this work, we utilize plain knowledge distillation, which is based on logits to avoid forgetting.

\paragraph{Bias removal methods:}
Various works~\cite{lucir, bic, fairness} have pointed out that class-imbalance between old and new classes creates a bias in the class weight vectors in the last linear layer, due to which the network predictions are biased towards new classes.
To rectify this bias, Wu~\etal~\cite{bic} trained an extra bias-correction layer using the validation set, Belouadah~\etal~\cite{Belouadah_2019_ICCV} proposed to rectify the final activations using the statistics of the old task predictions, Zhao~\etal~\cite{fairness} adjusted the norm of new class-weight vectors to those of the old class-weight vectors, and 
Hou~\etal~\cite{lucir} applied cosine normalization in the last layer.
The focus of these works is limited to the bias in the last layer, but ultimately catastrophic forgetting is an issue that affects the entire network: class imbalance causes the model to overfit to the new task, deteriorating the performance on the old ones. 
Some works \cite{eeil, unlabeled} also fine-tune the model to avoid overfitting to the current task. We propose a learning system that resolves this bias without the need of any post-processing, by fixing the underlying issues; see Section \ref{sec:compo_loss}.

\section{Class-Incremental Learning}

\subsection{Problem Definition}
The objective of class-incremental learning (class-IL) is to learn a unified classifier from a sequence of data from different classes. Data arrives incrementally as a batch of per-class sets $\mathcal{X}$ \ie~($X^1, X^2,...,X^t$), where $X^y$ contains all images from class $y$. Learning from a batch of classes can be considered as a task $\mathcal{T}$. 
At each incremental step, the data for the new task $\mathcal{T}_i$ arrives, which contains samples of the new set of classes. 
At each step, complete data is only available for new classes $\mathcal{X}$~\ie~($X^{s+1},...,X^t$). Only a small amount of exemplar data $\mathcal{P}_{old}$ \ie~($P^1,...,P^s$) from previous classes \ie~($X^1,...,X^s$) is retained in a memory buffer of limited size. The model is expected to classify all the classes seen so far. 

The problem definition with strictly separated batches may appear a bit specific. In many practical applications, the data will arrive in a more mixed-up fashion. However, this strict protocol allows the comparison of techniques and it covers the key issues with class-incremental learning. Improvements on this protocol also serve less strict applied settings. 

\subsection{Evaluation Metrics for Class-IL}
\label{sec:classil_metrics}
Class-IL models are evaluated using three metrics: average incremental accuracy, forgetting rate and feature retention.
After each incremental step, all classes seen so far are evaluated using the latest model. After $N$ incremental tasks, the accuracy $\mathcal{A}_n$  over all ($N+1$) steps is averaged and reported. It is termed as \textit{average incremental accuracy} (\textbf{Avg Acc}), introduced by Rebuffi~\etal~\cite{icarl}.
We also evaluate the\textit{ forgetting rate} $\mathbfcal{F}$ proposed by Liu~\etal~\cite{mnemonics}. The forgetting rate measures the performance drop on the first task. It is the accuracy difference on the classes of the first task $X^{1:s}_{test}$, using $\Theta_0$ and $\Theta_N$. Therefore, it is independent of the absolute performance on the initial task $\mathcal{T}_0$. 
We introduce another metric, referred as \cfr, to measure retention in the feature extractor $\phi(\cdot)$. It measures how much information is retained in the feature extractor while learning the tasks incrementally as compared to a jointly trained model. To measure \cfr: after the final incremental step, parameters of the feature extractor are frozen and the last linear layer is learned using all the data from all the classes. \cfr~is the accuracy difference between this model and a model where the whole network is trained on all the classes with complete data access. 

\begin{figure*}[t]
\centering
\begin{subfigure}[b]{.5\textwidth}
 \begin{tabular}{cc}
  \centering
  \includegraphics[width=0.9\linewidth]{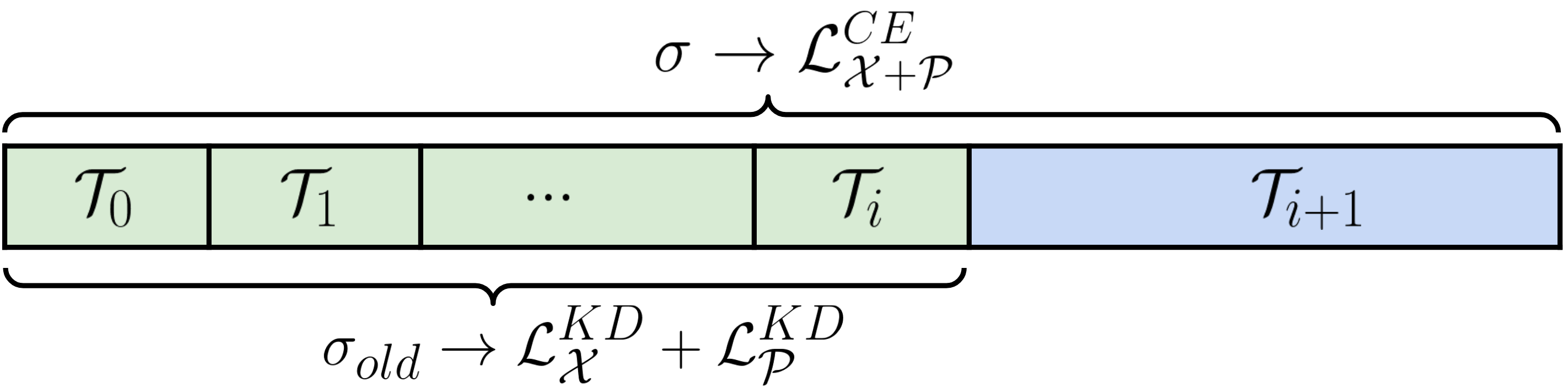}
  \end{tabular}
  \caption{}
  \label{fig:comb_softmax}
\end{subfigure}%
\begin{subfigure}[b]{.5\textwidth}
 \begin{tabular}{cc}
  \centering
  \includegraphics[width=0.9\linewidth]{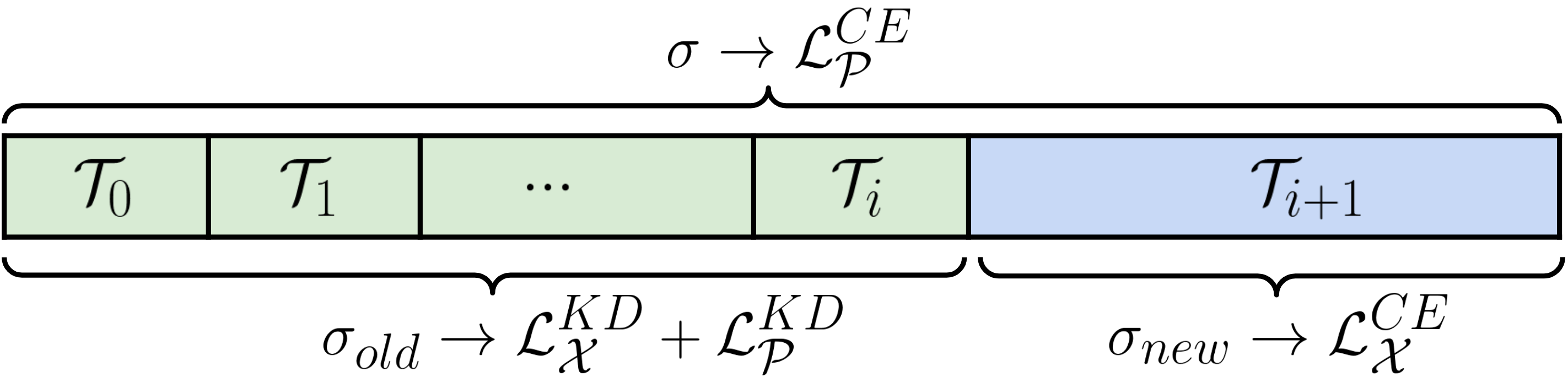}
  \end{tabular}
  \caption{}
  \label{fig:sep_softmax}
\end{subfigure}
\caption{The comparison between a (a) standard loss system and our proposed (b) compositional loss system (right). $\sigma$ shows the softmax function span over all the network output logits. $\sigma_{old}$ and $\sigma_{new}$ shows softmax span over the set of old and new class logits respectively. }
\label{fig:ccil_vs_icarl}
\end{figure*}

\begin{algorithm}[!b]
\SetAlgoLined
\DontPrintSemicolon
\Indentp{0.5em}
\Indm
\KwIn{ $ \mathcal{X} = (X^{s+1},...,X^t), \mathcal{P}^s = (P_1,...,P_{s})$  //  new classes data, old exemplar sets}
\KwIn{$K, \Theta^s, \hat{\Theta}^s$  \hspace{0.2cm} //  memory size, current model, frozen current model}
\KwOut{$\Theta^t$  \hspace{0.2cm}    // model trained on $t$ classes} 
\Indp
$\textit{m} \gets K/t$ \hspace{0.2cm} // number of exemplars per class\\
$\Theta^t \gets \Theta^s$  \hspace{0.2cm} // add output nodes for new classes \\
$\mathcal{P} \gets \text{UpdateExemplarSets}(\mathcal{X}; \mathcal{P}^s, m, \Theta^s)$  \label{alg1:new_exem} \\

\For(\hspace{100px}// {\textbf{update for mini-batch data in} $\mathcal{X}$} ){$(x,y)\in \mathcal{X}$}{  \label{alg1:intra_task_start}

$o = \Theta^t(x)$  \hspace{1cm}  // $o=\{o_{old}, o_{new}\}$\\   
softmax over new class logits $ \sigma_{new}(o_{new})$  \\
compute classification loss $\mathcal{L}^{CE}_{\mathcal{X}}$ (Eq. \ref{eq:cls_loss_x}) \\
softmax over old class logits $\sigma_{old}(o_{old})$  \\
 compute distillation loss $\mathcal{L}_\mathcal{X}^{KD}$  (Eq. \ref{eq:dist_loss_x})\\
 \label{alg1:intra_task_end}
\BlankLine
 \textbf{load a mini-batch from exemplars set} $(x',y')\sim \mathcal{P}$  \label{alg1:inter_task_start} \\
$o' = \Theta^t(x')$    \\
softmax over all logits $\sigma(o')$  \label{alg1:comb_sigma}  \\
compute classification loss $\mathcal{L}^{CE}_{\mathcal{P}}$ (Eq. \ref{eq:cls_loss_p}) \\
compute distillation loss $\mathcal{L}^{KD}_{\mathcal{P}}$ (Eq. \ref{eq:dist_loss_p}) \\
\BlankLine
$\mathcal{L} = (\mathcal{L}^{CE}_{\mathcal{X}} + \mathcal{L}^{CE}_{\mathcal{P}}) + \lambda*(\mathcal{L}^{KD}_{\mathcal{X}} + \mathcal{L}^{KD}_{\mathcal{P}})$ \label{alg1:final_loss}
}
\caption{CCIL: \text{IncrementalStep}} \label{alg1}
\end{algorithm}


\subsection{A Basic Class-IL Framework}
The network model $\Theta$ consists of a feature extractor $\phi(\cdot)$ and a fully-connected layer $fc(\cdot)$ for classification. Similar to a standard multi-class classifier, the output logits $o$ are processed through a softmax activation function $\sigma(\cdot)$ before cross-entropy loss $\mathcal{L}^{CE}$ is evaluated corresponding to the correct class.
For the initial base task $\mathcal{T}_0$, the model $\Theta^s$ learns a standard classifier for the first ($y \in y[1:s]$) classes. 
In the incremental step, the $fc$ layer is adapted to learn new classes ($y \in y[s+1:t]$) by adding new output nodes, whereas the other part of the network remains unchanged, resulting into a new model $\Theta^t$. The three main elements of class-IL are set up as follows. 

\paragraph{Exemplar selection:}
We compile the exemplar set by randomly selecting an equal number of samples ($m$) for each class. The samples are sorted in ascending order according to the distance from the mean of the feature vectors $\mu_i$ for each class separately.
Since the size of the limited memory is fixed ($K$), some samples of old classes are removed to accommodate exemplars from new classes. Samples with larger distances to the mean vector are removed first. Detailed steps are shown in Algorithm~\ref{alg2}.

\paragraph{Forgetting constraint:}
Our model uses knowledge distillation as the forgetting constraint. Knowledge distillation penalizes the change with respect to the output of the old model ($\Theta^s$) using KL-divergence, thus preserving the network's knowledge about the old classes. The distillation loss ($\mathcal{L}^{KD}$) is computed for the exemplar sets ($\mathcal{P}$) as well as for samples from the new classes ($\mathcal{X}$). The final loss for our CCIL model is a combination of cross-entropy loss $\mathcal{L}^{CE}$ for classification and distillation loss $\mathcal{L}^{KD}$ for mitigating catastrophic forgetting as shown in Algorithm~\ref{alg1}-Line~\ref{alg1:final_loss}.

\paragraph{Learning system:}
We propose a new compositional learning system which addresses the weight-bias issue in class-IL. The proposed loss isolates inter-task and intra-task learning for a balanced processing of data by appropriately normalizing the output logits. The task-agnostic parts are shared to yield improved efficiency. The details are presented in the next section.




\section{Compositional Learning System} \label{sec:compo_loss}


For each gradient update, the \ilkd model receives data in separate batches from the set of new classes $\mathcal{X}$ and the set of exemplars $\mathcal{P}$. $\mathcal{P}$ is the updated exemplar set which also includes equal size of exemplars from the current new classes. (see Algorithm \ref{alg1}-Line \ref{alg1:new_exem})
Instead of merging the batches, we propose to compute two separate losses for $\mathcal{X}$ and $\mathcal{P}$ mini-batches: 
\begin{align}
    \mathcal{L}_{\mathcal{X}} = \mathcal{L}^{CE}_{\mathcal{X}} +  \lambda*\mathcal{L}^{KD}_{\mathcal{X}}\\    
    \mathcal{L}_{\mathcal{P}} = \mathcal{L}^{CE}_{\mathcal{P}} +  \lambda*\mathcal{L}^{KD}_{\mathcal{P}}
\end{align}

\paragraph{Intra-task Learning:}
The classification loss for the new classes ($\mathcal{L}^{CE}_{\mathcal{X}}$) is computed using a dedicated softmax function $\sigma_{new}$ comprising logits of new classes only (Figure \ref{fig:sep_softmax}) computed as: 
\begin{align}
    \mathcal{L}^{CE}_{\mathcal{X}} = -\sum_{i=s+1}^t y[i]\cdot \log(p_{new}[i]) \label{eq:cls_loss_x}
\end{align}
for $(x,y)\in \mathcal{X}$, where $p_{new} = \sigma_{new}(o_{new})$, $o = \Theta^t(x)$ and output logits comprise $o=\{o_{old}, o_{new}\}$.
This allows the classifier weights for the new classes to be learned independently of the previous classes - while sharing the feature extractor, thus effectively eliminating the weight bias.
Distillation loss ($\mathcal{L}^{KD}_{\mathcal{X}}$) is always computed using $\sigma_{old}$ (see Figure \ref{fig:sep_softmax}), since output of new network $p_{old} = \sigma_{old}(o_{old})$ are compared against the output of previous model $\hat{p}= \sigma_{old}(\hat{\Theta}^s(x))$ as:
\begin{align}
    \mathcal{L}^{KD}_{\mathcal{X}} = D_{KL}( {\hat{p}||p_{old}}) \label{eq:dist_loss_x}
\end{align}
In case of a unified softmax, the weights of the old classes are suppressed by the larger amount of new class samples during training.
A similar analysis has been shown by~\cite{sep_softmax} for a fine-tuning setup. 

\paragraph{Inter-task Learning:}
The separate softmax helps intra-task learning for the new classes, but this does not yet discriminate the new from the old classes. For inter-task learning, we plan a balanced interaction between the samples of old and new classes. We compile an exemplar set $\mathcal{P}$ which contains equal numbers of samples from all classes including old and new classes. However small, such exemplar set enables the model to capture the inter-task relationship through the loss $\mathcal{L}^{CE}_{\mathcal{P}}$, which uses a combined softmax function $\sigma$ evaluated on all classes (see Figure \ref{fig:sep_softmax}). 
\begin{align}
    \mathcal{L}^{CE}_{\mathcal{P}} = -\sum_{i=1}^t y'[i]\cdot \log(q[i])  \label{eq:cls_loss_p}
\end{align}
for $(x',y')\in \mathcal{P}$, where $q = \sigma(o')$ and $o' = \Theta^t(x')$. The distillation loss is computed similar to Eq. \ref{eq:dist_loss_x},
\begin{align}
    \mathcal{L}^{KD}_{\mathcal{P}} = D_{KL}( {\hat{q}||q_{old}}) \label{eq:dist_loss_p}
\end{align}
where $\hat{q} = \sigma_{old}(\hat{\Theta}^s(x'))$ and $q_{old} = \sigma_{old}(o'_{old})$. This exemplar set is compiled before learning the incremental task, contrary to previous works, where it is always compiled after the incremental step. Figure \ref{fig:ccil_vs_icarl} shows how the loss terms are calculated using a separate softmax function~\ref{fig:sep_softmax} and also compares it to the unified softmax~\ref{fig:comb_softmax} used in previous works. 

\paragraph{Transfer Learning:}
We observed that a separate softmax does not remove the bias completely. Another cause for unbalanced class-weight vectors, and catastrophic forgetting in general, is the change in the data distribution between different tasks. We hypothesize that the effect of this distribution shift in the training data is more harmful to the previous knowledge when the transfer learning from old to new classes is poor, resulting in strong alteration of the parameters of the network. 
We propose to reduce the learning rate for the incremental steps as a simple way to improve transfer learning and mitigate the adverse effect of distribution shift. This further helps reduce the weight bias. The reduced learning rate on incremental steps depends on the scale and relevance of features learned in the base task, therefore it is determined experimentally.
Although lowering the learning rate is a standard technique when fine-tuning a network on a new dataset, its importance is underestimated and often missing in incremental learning works. We experimentally show its importance in ablation studies (Section \ref{sec:ablation}).

\begin{algorithm}
\Indentp{0.5em}
\Indm 
    \KwIn{$\mathcal{X}, \mathcal{P}_{old} $  // new class data, old exemplar set} 
    \KwIn{$\Theta^s, m$ // old model, new exemplar size per class} 
    \KwOut{$\mathcal{P}_{new}$  // new Exemplar sets}

\Indp
\For{$i=1,...,s$}{ 
 $P_i \gets (p_1,...,p_m)$   // keep first m samples
 }
 \tcc{add new class exemplars}
 \For{$i=s+1,...,t$}{
 $P_i \gets (p_1,...,p_m)\subset X^i )$   // randomly pick m samples \\
 $\mu_i \gets \frac{1}{m}\sum_{j=1}^m \phi(p_j)$ // mean feature \\ 
  \tcc{sort exemplars based on distance from $\mu_i$}
  \For{$k=1,...,m$}{
 $p_k \gets \argmin~|| \mu_i- \phi(p_k)||$}
 }
 \caption{UpdateExemplarSets} \label{alg2}
\end{algorithm}


\section{Improving Feature Representations for Incremental Learning}
Intuitively, poorly transferable embeddings will force the model to alter its parameters significantly in order to learn new concepts. This destroys the knowledge accumulated for the previous tasks. 
In this section, we explore this novel direction- aiming to learn robust representations that are transferable to a new task and effectively retain previous knowledge in class-IL. 
In particular, we study the detrimental effects of overfitting and loss of secondary class information. 
We find that: 1) both phenomena strongly correlate with catastrophic forgetting; 2) regularization methods can significantly improve robustness against forgetting, but only as long as they enhance the secondary class information of the learned model.


\begin{figure}[!t]
\includegraphics{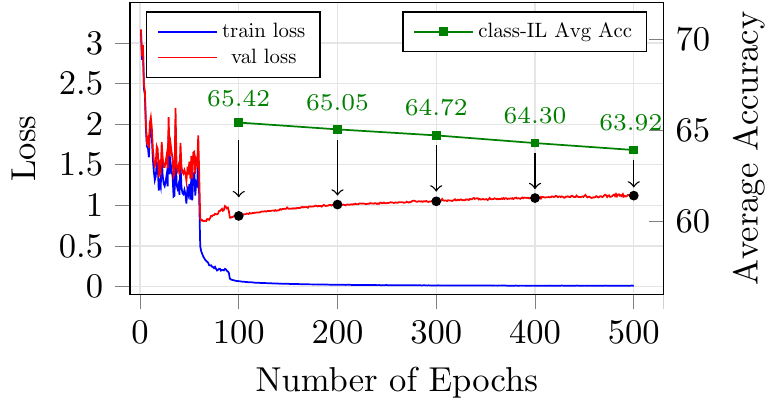}
\caption{The effect of overfitting on class-IL performance on the CIFAR-100 dataset. Figure shows the overfitting behavior on the initial base task. The validation loss (red curve) starts increasing monotonically after the $100^{th}$ epoch. The green curve shows the average incremental accuracy (right y-axis) for class-IL experiments performed over different snapshots at every $100^{th}$ epoch. }
\label{fig:overfit}
\end{figure}

\subsection{Measuring the Quality of Secondary Logits}
Secondary information captures semantic relationship between the target and non-target classes.
In literature, the term \textit{secondary information} is interchangeably used to denote the non-target and non-maximum scores of a classifier~\cite{kdingen_2019_AAAI}. 
Here, for evaluation purposes, the term denotes the non-maximum scores produced by the networks.
When applying the maximum operation to the scores predicted by a classifier, part of the information produced by the model is discarded.
For each individual sample this information represents the model's belief about the semantic nature of the image, in relation to the other classes.
It is important to learn this secondary information, such that the model can re-use it to learn new classes with least modification to previous concepts.
We argue that semantically similar classes should lie closer in the representation space as compared to the dissimilar classes since they share more features, and higher secondary information is an indicator of such an efficient non-redundant feature space. Appendix includes an analysis on feature representations to support this argument.

No proper annotations exist for secondary information, therefore we define a proxy evaluation objective, exploiting the  \textit{coarse}-labeling of the CIFAR-100 dataset, which partitions the 100 \textit{fine}-classes into 20 superclasses. The 5 classes belonging to each superclass are mostly semantically related, and have been previously used for evaluating secondary information~\cite{kdingen_2019_AAAI}.
As a proxy evaluation measure for secondary class information we propose to use the classification performance on the superclasses, restricting the network output to the non-maximum logits. We define two new metrics for this purpose: Secondary Superclass NLL and Secondary Superclass Accuracy.

\paragraph{Secondary Superclass-NLL (SS-NLL):}
Negative Log Likelihood is a commonly used cost function for classification, also known as \textit{Cross-Entropy Loss}.
Here we compute the NLL induced by the secondary (non-maximum) logits on the superclass classification problem. Given a set of superclasses $\mathcal{S}$, we can group the fine-grained classes into subsets $\mathcal{C}$ according to their coarse-label, and compute:
\begin{align}
    SS\text{-}NLL(x, y) = -\sum_{j\in\mathcal{S}}\bigg[\mathbb{1}_{\mathcal{C}_j}(y)\log\sum_{k\in\mathcal{C}_j} \hat{\sigma}_k\big(f(x)\big)\bigg],
\end{align}
where $\mathbb{1}_{\mathcal{C}_j}(y)$ is an indicator function which evaluates to 1 if the true class $y$ belongs to superclass $j$, $\hat{\sigma}$ is a softmax function over the secondary fine-logits (i.e. it suppresses the maximum logit). The network prediction (logits) is denoted as $f(x)$. A lower SS-NLL indicates better superclass classification and thus higher secondary information quality.
    
\paragraph{Secondary Superclass-Accuracy (SS-Acc):} Secondary superclass accuracy computes the percentage of correct superclass predictions. As for SS-NLL, the largest logit score is excluded from the prediction to focus the measure on the quality of secondary information.
Higher SS-Acc values indicate higher quality of the secondary information.

\begin{table}[!t]
\centering
\begin{tabular}{|l@{\hskip3pt}|c@{\hskip3pt}|c@{\hskip3pt}|c|@{\hskip3pt}c@{\hskip3pt}|}
    \hline
    \multirow{2}{*}{\textbf{Epoch}}& \textbf{SS-\hspace{5px}} & \textbf{SS-\hspace{5px}} & \multirow{2}{*}{$\mathbfcal{F}$ $\downarrow$} & \multirow{2}{*}{\cfr $\downarrow$}   \\ 
     & \textbf{NLL} $\downarrow$ & \textbf{Acc} $\uparrow$ & &  \\
        \hline
        \hline
      100 & 2.54    & 38.68 & 16.03  &  9.04  \\
      200 & 2.89    & 32.88 & 16.04  &  9.27 \\
      300 & 3.03    & 30.09 & 16.94  &  9.51 \\
      400 & 3.09    & 29.04 & 18.38  &  9.68 \\
      500 & 3.11    & 27.97 & 18.57  &  10.00 \\
        \hline
\end{tabular}
\caption{The effect of overfitting on class-IL performance and its correlation with secondary information, on the CIFAR-100 dataset. Table shows the performance of the snapshots taken at every $100^{th}$ epoch and the corresponding class-IL model.
SS-Acc decreases and SS-NLL increases as more overfitted models are evaluated. Forgetting rate $\mathbfcal{F}$ and feature retention metric \cfr~also correlate with overfitting. Results are averaged over 5 runs, standard deviation is reported in Appendix.}
\label{tab:overfit_forgetting}
\end{table}

\begin{table*}[t]
    \begin{center}
  \begin{tabular}{| @{\hskip10pt}l@{\hskip10pt}|@{\hskip10pt}c@{\hskip10pt}|@{\hskip10pt}c@{\hskip10pt}|c|c |c|c | c|}
   
    \hline
    
    \multirow{2}{*}{\textbf{Model}} &  \multicolumn{2}{c|}{\textbf{ Avg. Acc.}$\uparrow$} & \multicolumn{2}{c|}{\textbf{SS Metrics}}  &  \textbf{Forgetting} & \textbf{F. Retention} & \multirow{2}{*}{\textbf{ECE}$\downarrow$} \\
    
      & 5 tasks  & 10 tasks  & SS-NLL $\downarrow$ & SS-Acc. $\uparrow$  & $\mathbfcal{F}$ $\downarrow$  & \cfr $\downarrow$ & \\
  \hline
    \ilkd                            & 66.44    & 64.86  & 2.784    & 34.83  & 17.13 & 9.70 & 0.100\\
    \ilkd + SD                       & \textcolor{dgreen}{67.17}             & \textcolor{dgreen}{65.86}     & \textcolor{dgreen}{2.675}     & \textcolor{dgreen}{37.26}  & \textcolor{dgreen}{16.81} &  \textcolor{dgreen}{8.88} & \textcolor{dgreen}{0.094}\\
    
    \ilkd +  H-Aug                   & \textcolor{dgreen}{71.66}             & \textcolor{dgreen}{69.88}     & \textcolor{dgreen}{2.051}     & \textcolor{dgreen}{47.69} & \textcolor{dgreen}{13.37} & \textcolor{dgreen}{6.73} & \textcolor{dgreen}{0.018}\\
    
    \ilkd + LS                       & \textcolor{dred}{63.08}               & \textcolor{dred}{61.99}       & \textcolor{dred}{3.103}       & \textcolor{dred}{24.25} & \textcolor{dred}{18.79} & \textcolor{dred}{12.83} & \textcolor{dgreen}{0.049}  \\
    
    \ilkd + Mixup                    & \textcolor{dred}{62.31}               & \textcolor{dred}{57.75}       & \textcolor{dred}{2.791}       & \textcolor{dred}{31.57} &  \textcolor{dred}{24.56} &  \textcolor{dred}{16.01} & \textcolor{dgreen}{0.024} \\

    \hline
    \end{tabular}
    \end{center}
    \caption{Effect of regularization class-IL average accuracy, secondary information (on the first-task model), forgetting rate and feature retention (5 tasks), on CIFAR-100. All the values are averaged over 3 runs. $\downarrow$ and $\uparrow$ in the column headings indicate that lower and higher values are better respectively.
    Values that are better than the \ilkd baseline are marked in green whereas the worse ones are marked in red. SD:self-distillation, LS:label-smoothing, H-Aug:heavy data augmentation. Standard deviation in Appendix A.}
    \label{table:reg}
\end{table*}


\subsection{Forgetting starts before the incremental step}
\label{sec:overfitting}

In this section, we study how the quality of the representations learned during the initial base task correlates with incremental learning performance. 
We experimentally show how a decline in quality of the learned features - measured as overfitting and loss of secondary information - leads to higher catastrophic forgetting, motivating our following search for a suitable regularizer.

\paragraph{Experiment details:} We set up a standard class-IL experiment (with 5 incremental tasks) on CIFAR-100, using a ResNet-32 model.
The initial base network is trained for up to 500 epochs. We employ a SGD optimizer with a base learning of 1e-1, weight decay of 5e-4 and momentum 0.9. We use a step learning rate schedule, where the learning rate is divided by 10 at $60^{th}$ and $90^{th}$ epochs.

\paragraph{Analysis:} Figure~\ref{fig:overfit} shows that the validation loss (red curve) starts increasing after about 100 epochs, showing an overfitting effect.
Thereafter, we perform five different class-IL experiments, each based on a different snapshot of the base network (every $100^{th}$ epoch).
As the validation loss of the snapshot increases, incremental learning performance of the corresponding class-IL model drops (green curve), and both forgetting rate ($\mathbfcal{F}$) and feature retention metric (\cfr) worsen (Table~\ref{tab:overfit_forgetting}). The worsening \cfr~metric indicates that the issue is rooted in the feature representations, and cannot be mitigated by acting on the last layer bias.
Along with these metrics, we observe that overfitting causes the quality of secondary information to deteriorate (SS-Acc decreases and the SS-NLL increases, Table~\ref{tab:overfit_forgetting}).
This loss of secondary information could also be linked to increasing overconfidence of the network, measured as Expected Calibration Error (ECE) \cite{pmlr-v70-guo17a} (details in Appendix A).

These results indicate that: 
1) the quality of the features learned during the first base task influences the performance of the class-IL model, and as such it should be expressly addressed.
2) secondary information can be considered as an indicator of the features' quality and their fitness for incremental learning.
In the next section we will show experimental evidence in support of these hypotheses.

\subsection{Analyzing Catastrophic Forgetting with Regularization}
\label{sec:regularization}
Having established a link between early feature quality and catastrophic forgetting, we hypothesize that the application of adequate regularization techniques can improve model performance on the task at hand. We apply four common regularization techniques to our CCIL model: self-distillation~\cite{bornagain_nn}, data-augmentation (including cropping, cutout~\cite{cutout} and an extended set of AutoAugment~\cite{Cubuk_2019_CVPR} policies), label smoothing~\cite{label_smoothing}, and mixup~\cite{mixup}. All these regularizers have been shown to improve generalization on the held-out validation data. We report details about the application of said regularization methods in Appendix A.

\paragraph{Self-distillation}
Self-distillation \cite{bornagain_nn, sd_hilbert} is a form of knowledge distillation in which the teacher and student networks have the same architecture. It can be applied iteratively, in generations: at each generation a copy of the current student becomes the new teacher, with proven positive effects on generalization.

\paragraph{Data augmentation}
Augmentation is one of the most widespread regularization techniques for neural networks, especially in computer vision. A well designed data augmentation routine is key to obtaining good results on the held-out dataset. We sample randomly from a pool of augmentation policies which contain pairs of different geometric and color transformations, similarly to~\cite{Cubuk_2019_CVPR}.

\paragraph{Label smoothing}
Label smoothing~\cite{label_smoothing} acts on the cross-entropy loss for classification by interpolating the one-hot labels with a uniform distribution over the possible classes. This technique has been shown to improve generalization and reducing overconfidence of classification models~\cite{label_smoothing}. 

\paragraph{Mixup}
Mixup~\cite{mixup} is an operation that generates training samples for classification by linearly combining pairs of existing samples - image and label. 
Mixup has successfully been used as a form of data augmentation in image classification, improving generalization and calibration~\cite{mixup, thulasidasanmixup}. 


\begin{figure*}[!t]
\centering
\begin{subfigure}[b]{.3\textwidth}
 \begin{tabular}{cc}
\centering
\includegraphics[width = 1.9in]{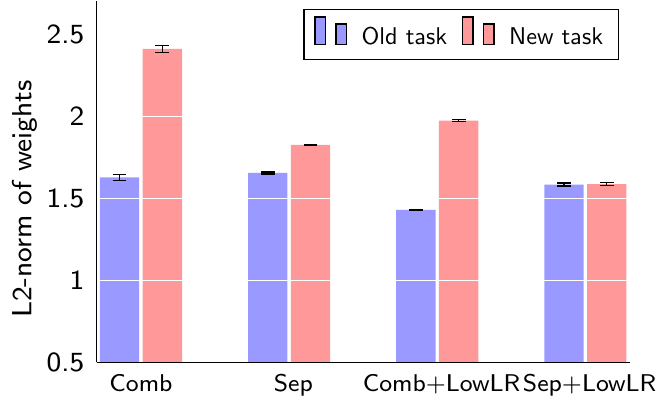} 
\end{tabular}
  \caption{}
  \label{fig:weight_norm_wo_kd}
\end{subfigure}%
\begin{subfigure}[b]{.3\textwidth}
 \begin{tabular}{cc}
\centering
\includegraphics[width = 1.9in]{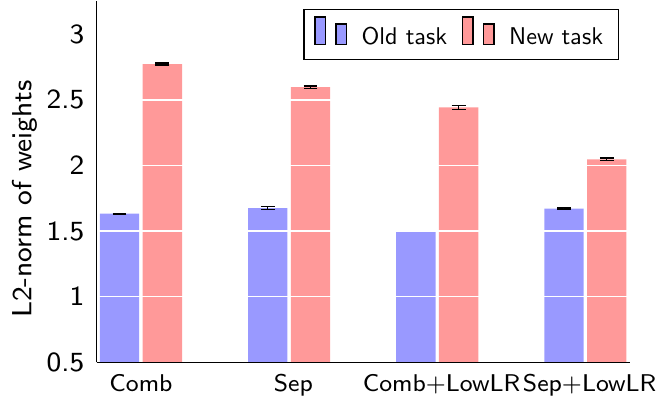} 
\end{tabular}
  \caption{}
  \label{fig:weight_norm_w_kd}
\end{subfigure}%
\begin{subfigure}[b]{.35\textwidth}
 \begin{tabular}{|c|c|c|} 
 \hline
 \multirow{2}{*}{\textbf{Operations}}   & \textbf{Avg Acc} $\uparrow$  & \textbf{Avg Acc} $\uparrow$ \\ [0.5ex] 
 & \textbf{w/o KD} & \textbf{w/ KD}\\
 \hline
 \hline

    Comb                        & {47.97}           & 52.71 \\ 
    \hline
    Sep                         & {52.86}           & 60.85 \\
      \hline
   Comb+LowLR                   & 52.79             & 54.54        \\
      \hline
    Sep+LowLR                   & \textbf{58.60}     & \textbf{64.79}        \\
    \hline
    \end{tabular}
    \vspace{10px}
     \caption{}
    \label{tab:weight_norm}
\end{subfigure}
\caption{ (a) \& (b) compares the average $L_2$ norm of the classification weight vectors for old and new classes for class-IL experiments without (w/o) and with (w/) KD respectively. We evaluate standard combined softmax (Comb) against proposed separate softmax (Sep) and we assess the effect of reduced learning rate (LowLR). (c) contains the corresponding class-IL results without distillation (w/o KD) and with distillation (w/ KD) in terms of average incremental accuracy. All experiments use the linear classification layer. Results shown on CIFAR-100 for 5-task experiments. } 
\label{fig:weight_norm_fig_tab}
\end{figure*}

\paragraph{Analysis} \hspace{5pt}
We analyse above discussed metrics for each of these regularization techniques. Table~\ref{table:reg} shows the Average Accuracy after finishing the last incremental step, secondary information quality of the first task model, forgetting rate, feature retention (Section~\ref{sec:classil_metrics}) and expected calibration error~\cite{pmlr-v70-guo17a}.
We can divide the regularization methods into two groups: the ones which improve class-IL performance (self-distillation, augmentation) and the ones which harm it (label smoothing, mixup). 
The first group also shows consistent improvements in secondary information and reduction in forgetting, with augmentation performing the best across all metrics - by a significant margin. 
In the second group, label smoothing harms secondary information the most. It has been observed that label smoothing encourages representations to be closer to their respective class centroid and equidistant to the other class centroids~\cite{NIPS2019_8717}, and this comes at the expense of inter-class sample relationships, i.e., secondary information.
Mixup also harms the quality of secondary information: we believe this is because it artificially forces arbitrary distances between classes, which modifies the natural output distribution - similarly to label smoothing. 
Interestingly, all regularizers improve network calibration, but ECE is not a good indicator of class-IL performance, unlike secondary information, shown in Table~\ref{table:reg}.

In summary, label smoothing and mixup - despite their proven regularization effects - harm secondary class information and have clear negative consequences for class-incremental learning.
On the other hand, regularization methods that enhance secondary class information (self distillation and data augmentation) boost the average incremental accuracy.
Analogously to the analysis of Section~\ref{sec:overfitting} we show that the quality of secondary information negatively correlates to the forgetting rate (Table~\ref{table:reg}), further indicating the importance of secondary class information.

\section{Results}

\subsection{Training Details}

\paragraph{Datasets} We conduct experiments on CIFAR100~\cite{Krizhevsky09learningmultiple}, ImageNet-100 Subset~\cite{imagenet_cvpr09} and full ImageNet datasets. CIFAR-100 contains 60K images from 100 classes of size $32\times32$, with 50K images for training and 10K for evaluation. The ImageNet-100 dataset has 100 randomly sampled classes (using Numpy seed:1993) from ImageNet.
The base \ilkd model uses default data augmentation including random cropping and horizontal flipping for CIFAR-100, and resized-random cropping and horizontal flipping for ImageNet datasets. All the randomization seeds are selected following the experiments in previous works~\cite{lucir, mnemonics}.

\paragraph{Benchmark protocol} We follow the protocol used in previous works \cite{lucir, mnemonics}. The protocol involves learning of 1 initial base task followed by $N$ incremental tasks. We evaluate with two incremental settings: where the model learns $N=5$ and $N=10$ incremental tasks. For CIFAR-100 and ImageNet-100, 50 classes are selected as the base classes for the initial task and the remaining classes are equally divided over the incremental steps. A similar format is followed for ImageNet with 500 base classes. Exemplar memory size is set to $K=2k$ for 100 class datasets and $K=20k$ for the full ImageNet dataset.

\paragraph{Implementation details} We use a 32-layer ResNet \cite{He_2016_CVPR} for CIFAR-100 dataset, and a 18-layer ResNet for ImageNet-100 and ImageNet datasets. The last layer is cosine normalized following the recommendations of \cite{lucir}. We will publish the code after acceptance. Hyperparameter details are included in the Appendix. 




\begin{table*}[!t]
\begin{center}
 \begin{tabular}{|c || c | c || c | c || c || c || c | c  || c || c |} 
 \hline
\multirow{2}{*}{\textbf{Method}} & \multicolumn{2}{c||}{\textbf{Layer}} &  \multicolumn{2}{c||}{\textbf{Softmax}} & {\textbf{Low}} & \multirow{2}{*}{\textbf{AW}} &  \multicolumn{2}{c||}{\textbf{Classifier}}  & \multirow{2}{*}{\textbf{KD}} & \multirow{2}{*}{\textbf{Avg Acc}}\\
 & Cos & Dot & Sep &  Comb  & {\textbf{LR}} &   & NME &  CNN  &  & \\ [0.5ex] 
 \hline\hline
Comb         &        & \yo   &        &  \yo  &         &        &   &  \yo    &      &   47.97   \\
 \hline
iCaRL       &        & \yo   &        &  \yo  &         &        &  \yo  &    &  \yo  &   56.50   \\
\hline
iCaRL++  &  \yo   &       &        &  \yo  &         &  \yo        &  & \yo  &  \yo  &   59.78       \\
\hline
CCIL  &  \yo   &       & \cellcolor{blue!15}\byo    &       &  \cellcolor{blue!15}\byo    &  \yo   &  &  \yo  &  \yo  &   66.44    \\
\hline
\end{tabular}
\end{center}
\caption{Drawing parallels between iCaRL and our proposed model. Average accuracy is reported for 5-task class-IL experiments on CIFAR-100 dataset. Last row highlights our proposed changes. All methods use random exemplar selection as used in this work, Dot: linear layer, KD: knowledge distillation, NME: nearest-mean-of-exemplars (used in~\cite{icarl})}
\label{tab:icarl_vs_ccil}
\end{table*}

\begin{table*}
\centering
 \begin{tabular}{  l@{\hskip 0.4in} c@{\hskip 0.15in}|@{\hskip 0.1in}c@{\hskip 0.4in}c@{\hskip 0.18in}|@{\hskip 0in}c@{\hskip 0.4in}c@{\hskip 0.15in}|@{\hskip 0.15in}c} 
 \hline 
  \textbf{Method}   & \multicolumn{2}{c@{\hskip 0.6in}}{\textbf{CIFAR-100}}   & \multicolumn{2}{c@{\hskip 0.6in}}{\textbf{ImageNet-100}} & \multicolumn{2}{c}{\textbf{ImageNet}}  \\
  
    \textbf{No. of incremental tasks} $\to$     & \textbf{5 }  & \textbf{10 }  & \textbf{5 }  & \textbf{10 }  & \textbf{5 }  & \textbf{10 }\\
 [0.5ex] 
 \hline

iCaRL$^{*}$ \cite{icarl}         & 57.17          & 52.57    & 65.04     & 59.53     & 51.50      & 46.89 \\ 
BIC \cite{bic}                   & 59.36          & 54.20    & 70.07     & 64.96     & 62.65     & 58.72\\
WA \cite{fairness}               & 63.25          & 58.57    &   ---     & ---       & ---       & --- \\
LUCIR \cite{lucir}               & 63.12          & 60.14    & 70.47     & 68.09     & 64.34     & 61.28 \\
Mnemonics \cite{mnemonics}       & 63.34          & 62.28    & 72.58     & 71.37     & 64.54     & 63.01 \\
TPCIL \cite{tpcil}               & 65.34          & 63.58    & 76.27     & 74.81     & 64.89     & 62.88 \\
\hline
\ilkd (ours)                 & 66.44            & 64.86    & 77.99     & 75.99      & 67.53       & 65.61 \\
\ilkdsd (ours)               & \textbf{67.17}   & \textbf{65.86} & \textbf{79.44}  & \textbf{76.77}    & \textbf{68.04}  & \textbf{66.25}  \\
\hline
 Joint-training             & 74.12          & 73.80    &  84.72    & 84.67     & 69.72       & 69.75\\
\hline 
\end{tabular}

\caption{Comparing average incremental accuracy computed using different methods on CIFAR-100, ImageNet-100 and ImageNet dataset. *as reported in \cite{lucir} }
\label{table:sota_results}
\end{table*}

\subsection{Ablation Studies} \label{sec:ablation}

\paragraph{Elements of the compositional learning system}
We evaluate the contributions of each element in the proposed learning system by training multiple class-IL models featuring them. 
The incremental learning in these experiments is conducted in two settings - in a simple fine-tuning setup (without distillation), in order to single out the effects of the proposed changes and with knowledge distillation loss included. 
In Figure \ref{fig:weight_norm_wo_kd} \& \ref{fig:weight_norm_w_kd} we compare the average L2 norm of the class weight vectors for old and new classes after 5 incremental training steps, while in Figure \ref{tab:weight_norm} we provide the average accuracies of the respective models. We notice a major difference in the weight norms of old and new classes for the default combined softmax (Comb) setting (Figure  \ref{fig:comb_softmax}). Using separate-softmax (Sep) substantially reduces this difference and improves class-IL performance, but does not resolve the problem completely.
Lower learning rate (Comb+LowLR) also removes the bias and improves the performance, although to a lesser extent.
When both approaches are combined (Sep+Low-LR), this bias is largely resolved and the best class-IL results are produced.

\paragraph{Drawing parallels with iCaRL}
We compare different components of our \ilkd model with the first baseline approach (iCaRL) proposed by \cite{icarl}. Table \ref{tab:icarl_vs_ccil} summarizes these changes. 
We first isolate the contributions of some follow-up methods by creating another baseline as iCaRL++. It consists of a (1) cosine-normalized layer (cos) \cite{Gidaris_2018_CVPR, DBLP:conf/icann/LuoZXWRY18, lucir}, where the features and class-weight vectors in the final layer are normalized to lie in a high-dimensional sphere. It helps in removing the remaining weight bias during inference, and (2) adaptive weighting (AW), where the weight of the distillation loss increases with incremental steps. AW was previously introduced in \cite{lucir}, which helps in adaptive balancing of classification and distillation loss (more details are included in the Appendix).
The last row shows that replacing the combined-softmax (comb) with the proposed separate-softmax (sep) and reducing the learning rate (LowLR) yields a major improvement.

\subsection{Comparison to SOTA}
Results for CIFAR-100, ImageNet-100 and ImageNet datasets are shown in Table \ref{table:sota_results}. 
We report the upper bound `Joint-training', where at every incremental step all the data for the classes seen until then is accessible.
The simple \ilkd model compares favorably to previous results on all datasets, especially on larger datasets like ImageNet-1k. 
The regularized \ilkdsd closes the gap to joint training further and achieves state-of-the-art performance across all datasets. 
Since the \ilkd model is based only on simple components, we believe that the application of advanced methods for mitigating forgetting \cite{lucir, tpcil} and more informative exemplar selection \cite{mnemonics} can further improve the performance. 


\section{Conclusions}




We presented a straightforward class-incremental learning system that focuses on the essential components and already exceeds the state of the art without integrating sophisticated modules. This makes it a good base model for future research on advancing class-incremental learning. 

Moreover, we showed that countering catastrophic forgetting during the incremental step is not enough: the quality of the feature representation prior to the incremental step considerably determines the amount of forgetting.
This suggests that representation learning is a promising direction to maximize also incremental performance.
In this regard we showed that boosting secondary information is key to improve the transferability of features from old to new tasks without forgetting.




\appendix

\section{Experiment Details}

 This section includes details concerning experiments included in the paper: Adapting weighting (AW) details, hyperparameter details and standard deviation for all the results. We report all the evaluation metrics averaged over 3 run trials (unless mentioned otherwise) to capture the variance in the class-IL training process.

  \begin{table*}[!t]
\centering
\begin{tabular}{|l@{\hskip3pt}|c@{\hskip3pt}|c@{\hskip3pt}|c@{\hskip3pt}|c|c|c|}
\hline
\textbf{Epoch}& \textbf{SS-NLL} $\downarrow$ & \textbf{SS-Acc} $\uparrow$ & \textbf{Avg Acc} $\uparrow$ & $\mathbfcal{F}$ $\downarrow$ & \cfr $\downarrow$ & ECE\\
    \hline
    \hline
    100 & 2.54 $\pm$ 0.04  & 38.68 $\pm$ 0.89  &   65.42  $\pm$ 0.06    & 16.03  $\pm$ 0.36  & 9.04  $\pm$ 0.24 & 0.093$\pm$0.003\\
    200 & 2.89 $\pm$ 0.06  & 32.88 $\pm$ 0.59  &   65.05  $\pm$ 0.08    & 16.04  $\pm$ 0.26  & 9.27  $\pm$ 0.42 & 0.118$\pm$0.003\\
    300 & 3.03 $\pm$ 0.06  & 30.09 $\pm$ 0.53  &   64.72  $\pm$ 0.07    & 16.94  $\pm$ 0.61  & 9.51  $\pm$ 0.23 & 0.126$\pm$0.004\\
    400 & 3.09 $\pm$ 0.07  & 29.04 $\pm$ 0.68  &   64.3   $\pm$ 0.12    & 18.38  $\pm$ 0.19  & 9.68  $\pm$ 0.17 & 0.131$\pm$0.005\\
    500 & 3.11 $\pm$ 0.03  & 27.97 $\pm$ 0.54  &   62.92  $\pm$ 0.11    & 18.57  $\pm$ 0.39  & 10.00  $\pm$ 0.20 & 0.137$\pm$0.002\\
    \hline
\end{tabular}
\vspace{5px}
\caption{The effect of overfitting on class-IL performance and its correlation with secondary information. Table shows the performance of the network snapshots taken at every $100^{th}$ epoch.
Accuracy decreases and SS-NLL increases, both monotonically, as more severely overfitted models are evaluated. Forgetting rate $\mathbfcal{F}$ also correlates with overfitting. Results are computed over 5 runs.}
\label{tab:overfit_std}
\end{table*}

\begin{table*}[!t]
    \begin{center}
    \begin{tabular}{l  c|c  c|c c|c }
    \hline
    \multirow{2}{*}{\textbf{Model}} &   \multicolumn{2}{c}{\textbf{ Avg. Acc.} $\uparrow$} & Forgetting (5 tasks) & Retention &  \multicolumn{2}{c}{\textbf{SS Metrics} (5 tasks)} \\ 
    
     \cmidrule{2-3} \cmidrule{4-5} \cmidrule{6-7} 
      & 5 tasks  & 10 tasks & $\mathbfcal{F}$ $\downarrow$  & \cfr $\downarrow$ & SS-NLL $\downarrow$ & SS-Acc. $\uparrow$ \\
  \hline
    \ilkd    & 66.44  $\pm$ 0.31     & 64.86 $\pm$ 0.40    & 17.13 $\pm$ 1.12    & 9.70 $\pm$ 0.15 &  2.784 $\pm$ 0.014 & 34.83 $\pm$ 0.654\\
    
    \ilkd + SD                       & \textcolor{dgreen}{67.17} $\pm$ 0.14       & \textcolor{dgreen}{65.86} $\pm$ 0.29   & \textcolor{dgreen}{16.81} $\pm$ 0.25  &  \textcolor{dgreen}{8.88} $\pm$ 0.35 & \textcolor{dgreen}{2.675} $\pm$ 0.037 & \textcolor{dgreen}{37.26} $\pm$ 0.251\\
    
    \ilkd +  H-Aug                   & \textcolor{dgreen}{71.66} $\pm$ 0.23       & \textcolor{dgreen}{69.88} $\pm$ 0.36    & \textcolor{dgreen}{13.37}  $\pm$ 0.60 & \textcolor{dgreen}{6.73} $\pm$ 0.45 & \textcolor{dgreen}{2.051} $\pm$ 0.013 & \textcolor{dgreen}{47.69} $\pm$ 0.590\\
    
    \ilkd + LS                      & \textcolor{dred}{63.08 } $\pm$ 0.21         & \textcolor{dred}{61.99}  $\pm$ 0.30     &  \textcolor{dred}{18.79} $\pm$ 0.29 & \textcolor{dred}{12.83}  $\pm$ 0.41 & \textcolor{dred}{3.103} $\pm$ 0.013 & \textcolor{dred}{24.25} $\pm$ 0.278\\
    
    \ilkd + Mixup                  & \textcolor{dred}{62.31 }  $\pm$ 0.46         & \textcolor{dred}{57.75} $\pm$ 1.64      &  \textcolor{dred}{24.56}  $\pm$ 2.52 &  \textcolor{dred}{16.01} $\pm$ 0.16 & \textcolor{dred}{2.791} $\pm$ 0.006 & \textcolor{dred}{31.57} $\pm$ 0.256\\
    \hline
    \end{tabular}
    \end{center}
    \caption{Effect of regularization on class-IL performance and secondary information. All the metrics are evaluated on the network trained on the first task. $\downarrow$ and $\uparrow$ in the column headings indicate that lower and higher values are better respectively. Values that are better than our baseline method (\ilkd) are marked in green whereas the worse ones are marked in red. SD:self-distillation, LS:label-smoothing.}
    \label{table:app:reg:il}
\end{table*}

\subsection{Adaptive Weighting (AW)}

In each incremental step, training a network comprises a classification loss and a distillation loss to preserve knowledge about previous classes. Our baseline contains an adaptive weighting function $\lambda$ (similar to~\cite{lucir}) between two losses:  
\begin{align}
   \lambda = \lambda_{base}\bigg(\frac{C_n+C_o}{C_n}\bigg)^{2/3} 
\end{align}

,where $C_n$ denotes number of new classes, $C_o$ denotes number of old classes, $\lambda_{base}$ is fixed constant for each method. It dynamically increases weightage on preserving old knowledge as incremental training continues. It improves the baseline model by $0.45\%$ for 5 task experiment on CIFAR-100. $\lambda_{base}=5$ is set for CIFAR-100, $\lambda_{base}=20$ for ImageNet-100 and $\lambda_{base}=600$ for ImageNet.  

\subsection{Experiment Details}

\paragraph{Dataset:}
CIFAR-100 classes are shuffled using a fixed seed (Numpy~\cite{5725236} seed:1993) across all methods for fair comparison.
The ImageNet-100 dataset has 100 randomly sampled classes (using Numpy seed:1993) from ImageNet and further shuffled (using Numpy seed:1993). It  contains around 128K images of size $224\times224$ for training and 5K images for evaluation. ImageNet-1k classes are also shuffled using a Numpy seed:1993. 

\paragraph{Optimizer:} On CIFAR-100, the base network is trained for 120 epochs using a cosine learning rate schedule, where the base learning rate is 1e-1. Subsequent $N$ tasks are trained for 240 epochs with a base learning rate of 1e-2.
The learning rate is decayed until 1e-4. We use a batch size of 100 for CIFAR-100 experiments.
Networks for CIFAR-100 dataset is optimized using the SGD optimizer with a momentum of 0.9 and weight decay of 5e-4. 

For ImageNet-100, the network is trained for 70 epochs using a step learning rate schedule, where the base learning rate is 1e-1 for the base task and 1e-2 for the subsequent $N$ tasks. The base learning rate is divided by 10 at \{30, 60\} epochs. 

For ImageNet, base task is trained for 70 epochs following a step learning rate, where the base learning is 1e-1. The base learning rate is divided by 10 at \{30, 60\} epochs.
The incremental task is trained for 40 epochs following a step learning rate, where the base learning rate starts from 1e-2. The base learning rate is divided by 10 at \{25, 35\} epochs. 
Networks for ImageNet datasets are optimized using the SGD optimizer with a momentum of 0.9 and weight decay of 1e-4. We use a batch size of 128 for both ImageNet datasets.



\subsection{Overfitting Experiment}

\paragraph{Results with standard deviation} Table~\ref{tab:overfit_std} shows class-IL performance using average accuracy and forgetting rate, and quality of secondary information using SS-NLL and SS-Acc for each class-IL runs using increasingly overfitted model snapshots. Average incremental accuracy and forgetting rate is computed for class-IL model trained over different snapshots (every $100^{th}$) from the above run. Table~\ref{tab:overfit_std} also shows expected calibration error (ECE) with standard deviation for different snapshots of the overfitted model. It shows that ECE monotonically increases with the number of training epochs.
Tables includes values averaged over 5 runs with respective standard deviation.


\subsection{Regularization}
\label{app:regularization}
All the regularizers are applied at base and all incremental steps, however major improvement is observed due its usage in the initial base task. 

\paragraph{Self-distillation}
In the experiments, self-distillation is conducted over 4 generations (optimized using validation performance) for CIFAR-100 and ImageNet-100 dataset, and over 2 generations for ImageNet dataset. In the beginning of each self-distillation generation, the network snapshot (student) becomes the teacher network and the student continues to train (fine-tuned) with a combination of classification and distillation loss. 

For CIFAR-100 experiments, the first base model is trained for 120 epochs following a cosine learning rate schedule, decaying from a learning rate 1e-1 to 1e-4. For self-distillation generations, the model is trained for 70 epochs with a decaying (cosine) learning rate from 1e-1 to 1e-3. All other optimizer settings are the same as the baseline model.

For ImageNet-100 experiments, first base model is trained for 70 epochs following a step learning rate schedule. For self-distillation generations, the model is trained for 30 epochs each where base learning rate is 1e-2 and it is divided by 10 at {10, 20} epochs.

For ImageNet experiments, the first base model is trained for 70 epochs following a step learning rate schedule. For self-distillation generations, the model is trained for 15 epochs each where base learning rate is 1e-2 and it is divided by 10 at {8, 12} epochs.

\begin{figure*}[!t]
    \centering
    \begin{subfigure}{0.40\textwidth}
        \centering
        \includegraphics[width=\textwidth]{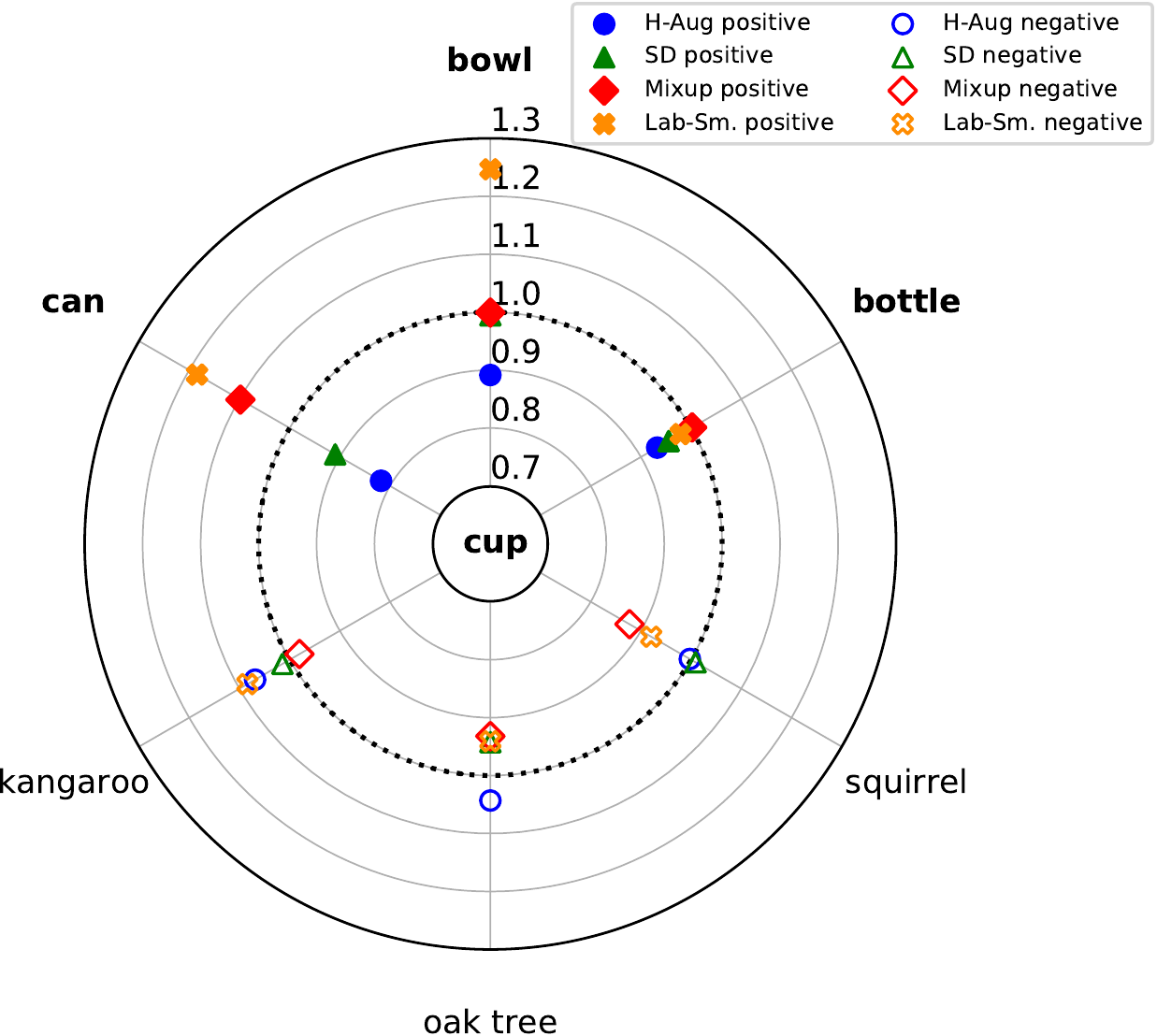}
        \caption{}
    \end{subfigure}%
    ~ 
    \begin{subfigure}{0.40\textwidth}
        \centering
        \includegraphics[width=\textwidth]{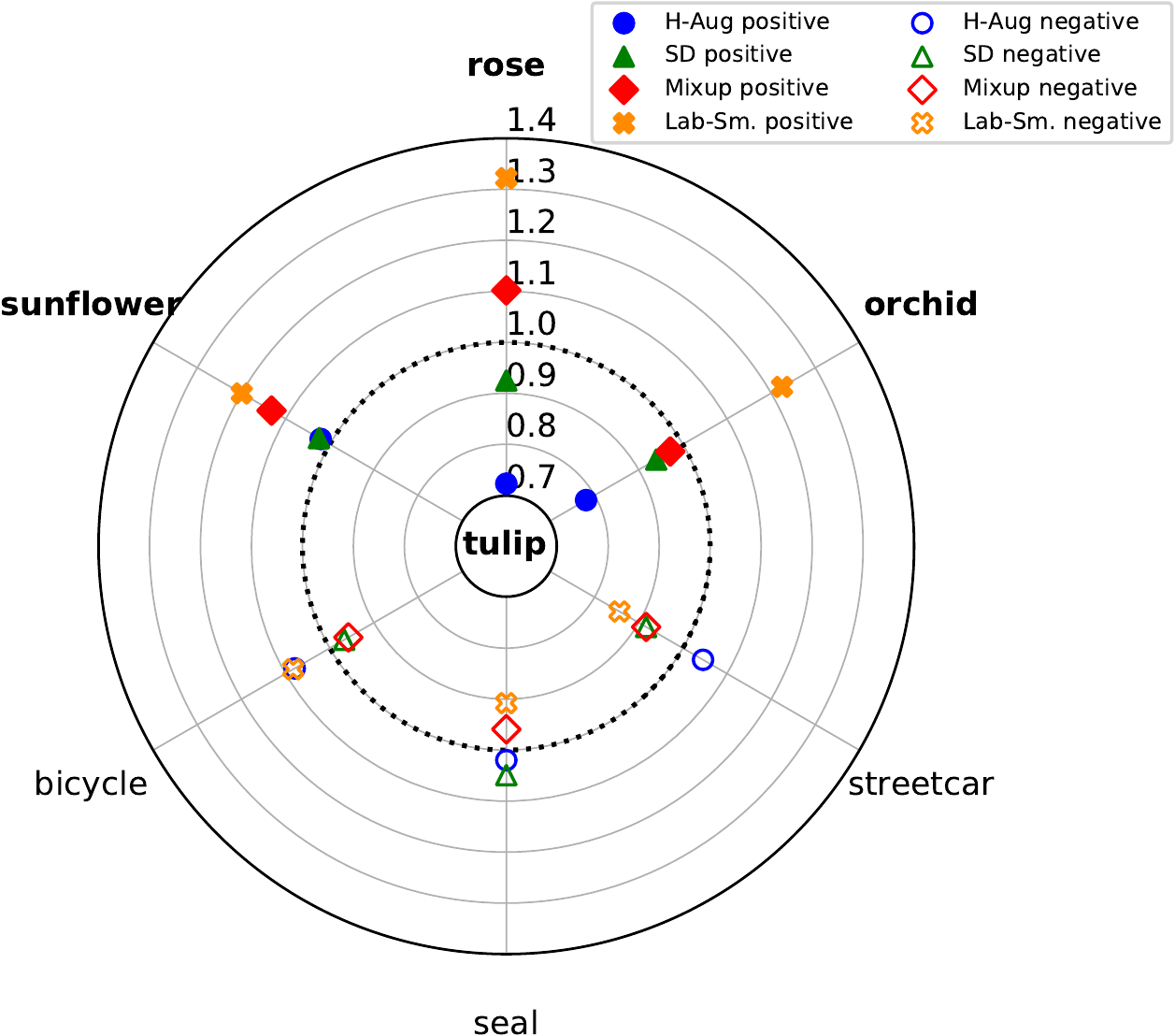}
        \caption{}
    \end{subfigure}
    \caption{Effect of regularizers on the distance between mean class representations. 
    The numbers shown in the plot are the ratios between the class means distances of each method and of the default \ilkd model.
    Similar classes are marked in \textbf{bold}.\textit{ Dotted} circle at 1.0 depicts distances between classes in the baseline \ilkd model and other distances are depicted relative to the baseline model. \textit{Positive} and \textit{negative} cases indicate similar and dissimilar classes respectively.  }   
    \label{fig:app:class_mean}
\end{figure*}

\paragraph{Results with standard deviations}
Table~\ref{table:app:reg:il} shows the effect of different regularization on the quality of secondary class information and the effect of different regularization on class-IL performance in terms of average incremental accuracy and forgetting rate. All experiments are conducted on CIFAR-100 dataset.

\section{Representations: Qualitative Analysis}

This section provides a qualitative analysis on the effect of different regularizers on the feature representations (penultimate-layer activations). We analyze the representations of the network trained on 50 classes (first task) of CIFAR-100 dataset using ResNet-32 network.

\subsection{Class-mean Representations}\label{sec:app:class_mean}

We argue that the classes which are semantically similar must be closer in the representation space as compared to the dissimilar classes since they share more features.
Based on this argument we analyze the effect of different regularization methods on the relative distances between class-mean representations. 
We utilize the fine- and coarse-label structure of the CIFAR-100 dataset to compare the effect on the distance between semantically similar and dissimilar classes relative to the default baseline model. Classes associated with the same coarse label or superclass are considered as similar classes, whereas dissimilar classes are picked from different superclasses. L2 distance is used as the distance metric.

Figure~\ref{fig:app:class_mean} show this qualitative analysis for two classes: \textit{cup} and \textit{tulip}. For example cup and can are semantically similar classes. When self-distillation and augmentation are used as regularizers, the relative distance reduces to 0.9 and 0.8 respectively, whereas when label-smoothing and mixup are applied, the relative distance increases to 1.2 and 1.1 respectively. Other similar classes follow a similar trend, whereas dissimilar pairs show an opposite behavior. 
Overall we find that regularizers: self-distillation and heavy data-augmentation reduce the relative distance between the similar classes (marked in bold) while not affecting or increasing distance between dissimilar classes. Whereas mixup and label smoothing increase the relative distance between similar classes and reduce the relative distance between dissimilar classes. We notice that these observations agree with the findings on secondary class information presented in the main paper. 

Earlier in the main paper, we argued that label-smoothing and mixup regularization deteriorate secondary class information since they dismantle the natural output distribution. This qualitative analysis supports our argument showing how they conversely hamper the distances between similar and dissimilar classes.



\section*{Acknowledgements} This study was supported by the German Federal Ministry of Education and Research via the project Deep-PTL.



\clearpage

\end{document}